\documentclass{article}

     \PassOptionsToPackage{numbers, compress}{natbib}
\usepackage{algorithm,algorithmic}
\usepackage{makecell}
\usepackage{amssymb, multirow, paralist, color,amsmath,amsthm}

\usepackage{enumitem}
\usepackage{textcase}
\usepackage[T1]{fontenc}
\usepackage{mathtools}
\usepackage{cleveref}
  
\usepackage{empheq}
\usepackage{color,tikz}
\definecolor{darkgreen}{rgb}{0.00,0.5,0.00}
\usepackage{fancybox}
\usepackage{url}

\definecolor{myteal}{RGB}{27,158,119}
\definecolor{myorange}{RGB}{217,95,2}
\definecolor{myred}{RGB}{231,41,138}
\definecolor{mypurple}{RGB}{152,78,163}
\definecolor{myblue}{rgb}{.9, .9, 1}
\definecolor{mygreen}{RGB}{0,100,0}
\definecolor{mycyan}{rgb}{0.88,1,1}
\definecolor{mydarkred}{RGB}{192,47,25}

\def \R {\mathbb{R}}

\def \w {\mathbf{w}}

\def \x {\mathbf{x}}

\def \x {\mathbf{x}}

\def \1 {\mathbf{1}}

\usepackage[skins]{tcolorbox}

\def \y {\mathbf{y}}

\def \P {\mathcal{P}}

\usepackage{cancel}
\usepackage{colortbl}
\usepackage[colorinlistoftodos,bordercolor=orange,backgroundcolor=orange!20,linecolor=orange,textsize=normalsize]{todonotes}

\usepackage{booktabs}
\usepackage{tablefootnote}

\newlength\mytemplen
\newsavebox\mytempbox

\makeatletter
\newcommand\mybluebox{%
	\@ifnextchar[%]
	{\@mybluebox}%
	{\@mybluebox[0pt]}}

\def\@mybluebox[#1]{%
	\@ifnextchar[%]
	{\@@mybluebox[#1]}%
	{\@@mybluebox[#1][0pt]}}

\def\@@mybluebox[#1][#2]#3{
	\sbox\mytempbox{#3}%
	\mytemplen\ht\mytempbox
	\advance\mytemplen #1\relax
	\ht\mytempbox\mytemplen
	\mytemplen\dp\mytempbox
	\advance\mytemplen #2\relax
	\dp\mytempbox\mytemplen
	\colorbox{myblue}{\hspace{1em}\usebox{\mytempbox}\hspace{1em}}}

\def \y {\mathbf{y}}

\def \x {\mathbf{x}}

\def \D {\mathcal{D}}

\def \w {\mathbf{w}}

\def \R {\mathbb{R}}

\usepackage{amssymb}% http://ctan.org/pkg/amssymb
\usepackage{pifont}% http://ctan.org/pkg/pifont

\def \P {\mathcal{P}}

     \usepackage[final]{neurips_2021}

\title{Deep AUC Maximization for Medical Image Classification: Challenges and  Opportunities}

\author{%
  Tianbao Yang \\
  Department of Computer Science\\
  The University of Iowa, Iowa City, IA 52242 \\
  \texttt{tianbao-yang@uiowa.edu} \\
}

\begin{document}

\maketitle

\begin{abstract}
In this extended abstract, we will present and discuss opportunities and challenges brought about by a new deep learning method  by AUC maximization (aka \underline{\bf D}eep \underline{\bf A}UC \underline{\bf M}aximization or {\bf DAM}) for medical image classification. Since AUC (aka area under ROC curve) is a standard performance measure for medical image classification, hence directly optimizing AUC could achieve a better performance for learning a deep neural network than minimizing a traditional loss function (e.g., cross-entropy loss). Recently, there emerges a trend of using deep AUC maximization for large-scale medical image classification. In this paper, we will discuss these recent results by highlighting  (i) the advancements brought by stochastic non-convex optimization algorithms for DAM; (ii) the promising results on various medical image classification problems. Then, we will discuss challenges and opportunities of DAM for medical image classification from three perspectives, feature learning,  large-scale optimization, and learning trustworthy AI models.  
\end{abstract}

\section{A Brief history of AUC Maximization}
AUC maximization has a history of almost two decades. During these two decades, there have been  a variety of methods for optimizing the AUC score. In order to better position deep AUC maximization, we start by introducing the formulations and objectives for AUC maximization. 

{\bf Definitions and Formulations.} Let $(\x, \y)\sim \P$ denote an input data and label pair following a distribution $\P$, where $\x\in\R^d$ and $\y\in\{1, -1\}$. Let $h_\w(\cdot):\R^d\rightarrow\R$ denote a predictive model (e.g., a deep neural network, a linear model).   The ROC curve is obtained by plotting the true positive rate (TPR) vs. the false positive rate (FPR) by varying the prediction threshold. AUC can be calculated as the Riemann integral of the function TPR vs FPR. However, the complex nature of computing Riemann integral makes it difficult to design optimization algorithms towards maximizing AUC for learning the model $h_\w(\cdot)$. The most popular approach is to use the probability interpretation of AUC~\cite{hanley82}, i.e., $\text{AUC}(h_\w(\cdot), \P) = \Pr(h_\w(\x)> h_\w(\x')|y=1, y'=-1)$, which means that AUC (on the population level) is equivalent to the probability that a randomly selected positive data is ranked higher than a randomly selected negative data by the predictive function,  where $(\x, y=1)$ is a random positive data and $(\x', y'=-1)$ is a random negative data. On a given dataset $\D=\D_+\cup \D_-$, where $\D_+=\{(\x_i, y_i)\in\D: y_i=1\}$ and $\D_-=\{(\x_i, y_i)\in\D: y_i=-1\}$, the empirical AUC score can be computed as 
\begin{align}\label{eqn:eAUC}
    \text{AUC}(h_\w(\cdot), \D) = \frac{1}{n_+}\frac{1}{n_-}\sum_{\x_i\in\D_+}\sum_{\x_i'\in\D_-}s(h_\w(\x_i) - h_\w(\x_i')), 
\end{align}
where $n_+=|\D_+|, n_-=|\D_-|$, $s(a)=1$  if $a>0$, $s(a)=1/2$ if $a=0$, and $s(a)=0$ if $a<0$. %The AUC maximization problem is then cast as an minimization problem: 
%\begin{align}\label{eqn:eAUC}
% \min_{\w} \frac{1}{n_+}\frac{1}{n_-}\sum_{\x_i\in\D_+}\sum_{\x_i'\in\D_-}\ell(h_\w(\x_i), h_\w(\x_i')), 
%\end{align}
%where $\ell(\cdot, \cdot)$ is a pairwise surrogate loss. 

% $=\left\{\begin{array}{lc}1 & a>b\\ 1/2 & a=b\\ 0 & a<b\end{array}\right.$

{\bf Full Batch based Methods (2000 - 2010).} Earlier works for AUC maximization use full batch based methods, which process all examples at each iteration.  To the best of our knowledge, the earliest work dates back to 2000~\cite{Herbrich1999d}, which considers the ordinal regression problem and proposes support vector machine (SVM) formulation in the sense of pairwise classification. Later, the AUC maximization problem has been tackled by~\cite{10.5555/3041838.3041945} using gradient-based methods, by \cite{ca3e551a365f44598f995ed69ca9d2bc} for learning decision-tree models, by \cite{10.5555/945365.964285} in the framework of boosting, by~\cite{10.1145/1102351.1102399} in the framework of structured SVM. Some works also proposed speed-up techniques by reducing the number of pairs~\cite{DBLP:conf/icml/HerschtalR04,Rakotomamonjy04supportvector} or by reducing the number of iterations~\cite{10.5555/2503308.2503358}.   Nevertheless,  these full batch based algorithms could suffer a quadratic time complexity in the worst-case or a super-linear (e.g. log-linear) time complexity per-iteration, which makes them not scalable to large datasets.

{\bf Online Methods (2010 - 2015).} To the best of our knowledge, Zhao et al.~\cite{Zhaoicml11} is the first work that considers AUC maximizing in the online learning fashion. To deal with large-scale data, they proposed to maintain a dynamic buffer to store some historical data for updating the model parameter. This online learning approach was also studied in several later works~\cite{10.5555/2968826.2968904,pmlr-v28-kar13}. In~\cite{10.5555/2968826.2968904}, the authors also proposed mini-batch stochastic gradient methods that update the model parameters based on all data pairs in the mini-batch. However, since these online methods do not consider all data pairs their error bound or regret bound depends on the size of the buffer or mini-batch (e.g., $1/\sqrt{B}$). Hence, these algorithms will not converge unless the buffer size of mini-batch size $B$ is infinitely large. Gao et al.~\cite{gao2013one} proposed an one-pass AUC maximization algorithm based on the pairwise square loss for learning a linear model, which optimizes an online version of AUC at each iteration that pairs each received data with all historical data. To avoid storing all historical data, they leverage the structure of the square loss and maintain and update mean and covariance matrix of data. %In more general context,  online pairwise learning has been considered in many works~\cite{}. 

{\bf Stochastic mini-batch based Methods (2016 - present).}  \cite{ying2016stochastic} is a milestone work for stochastic optimization of AUC. They restricted their attention to the pairwise square loss and proposed to transform the non-decomposable objective into a decomposable min-max optimization problem, which favors stochastic methods based on mini-batch of data without explicitly constructing the pairs.  Later on, the convergence of stochastic optimization of AUC based on the min-max formulation was improved in~\cite{liu2018fast,natole2018stochastic,Natole2019StochasticAO}. The min-max formulation also serves as the basis for most recent works on DAM discussed in next section. 
		
				\vspace*{-0.1in}
\section{DAM and Applications in Medical Image Classification}		
		\vspace*{-0.1in}

 Recently, there is a trend of DAM  based on large-scale stochastic optimization algorithms for solving the non-convex min-max formulation of the pairwise square loss or its variants. We briefly discuss the developments for large-scale non-convex min-max optimization and its application to AUC maximization and then present some successful empirical studies on medical image classification.  
 
{\bf Non-Convex Min-Max Optimization and Stochastic DAM.} Stochastic non-convex min-max optimization algorithms were first analyzed in~\cite{rafique2018non} with provable convergence guarantee. Thereafter, the theoretical developments have been the major topic of a wave of  studies~\cite{2020arXiv200713605I,DBLP:journals/corr/abs-2006-06889,DBLP:journals/corr/abs-2008-08170,lin2019gradient,luo2020stochastic,DBLP:conf/nips/Tran-DinhLN20,yan2020sharp}.  Liu et al.~\cite{liu2019stochastic} developed the first practical and provable stochastic algorithms for DAM based on the min-max formulation of the pairwise square loss function, which enjoy a fast convergence rate. Recently, DAM has been also studied in the framework of federated learning~\cite{DBLP:conf/icml/GuoLYSLY20,DBLP:conf/icml/YuanGXYY21}. 

{\bf DAM for Medical Image Classification.} To the best of our knowledge, \cite{DBLP:journals/corr/abs-2012-03173} is the first work that evaluates the performance of DAM on large-scale medical image data with hundreds of thousands images, which is two orders larger than that was used in earlier works, e.g.,~\cite{sulam2017maximizing}. They proposed a new objective for robust AUC maximization to alleviate the issues of the square loss, namely the sensitivity to noisy data and the adverse effect on easy data. The new loss function can be also cast into a min-max objective, to which all existing non-convex min-max optimization algorithms can be applied. It was shown to be more robust than the commonly used  square loss, while enjoying the same advantage in terms of large-scale stochastic optimization. They conducted extensive empirical studies of DAM on four difficult medical image classification tasks, including (i) classification of chest X-ray images for identifying many threatening diseases, (ii) classification of images of skin lesions for identifying melanoma, (iii) classification of mammogram for breast cancer screening, and (iv) classification of microscopic images for identifying tumor tissue. In Table~\ref{tab:meddata}, we summarize their results on various medical image datasets.  The percentage in the column "Improvements" shows the improvement over the baseline method that is trained by minimizing the standard cross-entropy loss, the column "Competition Results" shows their rank over all participating teams in competitions. The authors also released a library for DAM called LibAUC~\footnote{\url{www.libauc.org}}. 

\begin{table}[t]
\center\caption{Summary of DAM's performance on several medical image classification tasks from~\cite{DBLP:journals/corr/abs-2012-03173}. The CheXpert data was released in the Stanford CheXpert competition~\cite{chexpert}, the Melanoam data was released in the 2020 Kaggle Melanoma Competition~\cite{kaggle}. }\label{tab:meddata}
\scalebox{0.95}{
\begin{tabular}{cccccc}
\hline
\textbf{Dataset}  & Image Domain& \#pos/\#all & \# Training  & Improvements & Competition Results\\ \hline
CheXpert &Chest X-ray & 20.21\% &224,316 & 2\% & 1/150+\\
Melanoma &Skin Lesion  & 7.1\%& 46,131& 1\% & 33/3314\\
DDSM+   &Mammogram &13\%& 55,000& 1.5\%& NA\\ 
PatchCamelyon  & Microscopic & 1\%& 148,960&5\% & NA\\ \hline
\end{tabular}}
\end{table}

\section{Challenges and Opportunities}
In this section, we will discuss outstanding challenges and new opportunities of DAM for medical image classification.

\textbf{Feature Learning.} Feature learning is an important capability of deep learning for tackling un-structured image data. The current practice of DAM uses a two-stage approach: the first stage is to learn the encoder network by optimizing the traditional cross-entropy loss and the second stage is to fine tune the encoder network and to learn the classifier by DAM~\cite{DBLP:journals/corr/abs-2012-03173}.  It is still not fully understood why optimizing the AUC loss in an end-to-end fashion does not yield better feature representations, and it remains an open problem how to learn better encoder networks by using DAM. One direction is to improve the end-to-end learning paradigm that could enjoy the benefit of both feature learning and robust classifier learning.  
%In our preliminary study, we propose to optimize a compostional objective in the form of $A(\w- \alpha\nabla L_{\text{CE}}(\w))$, where $A$ denotes a min-max AUC loss, and demonstrate that this approach can learn much better feature representations and yield better prediction performance in terms of AUC score. We show  results on some benchmark datasets in Figure~\ref{}. For the benchmark datasets, please refer to~\cite{DBLP:journals/corr/abs-2012-03173}. 
Another direction is to consider self-supervised pre-training methods on large-scale unlabeled medical datasets. This approach was recently explored in~\cite{DBLP:journals/corr/abs-2010-05352,DBLP:journals/corr/abs-2010-00747,azizi2021big}. But its success on downstream tasks of using DAM remains to be demonstrated. In this direction, we could consider pre-training on much larger medical datasets than those used in existing studies and demonstrate the performance of DAM on multiple downstream medical image classification tasks. 

\textbf{Large-scale Optimization.} Although large-scale optimization algorithms for DAM have been developed, there are still many open problems to be addressed. Below, we will list several important questions. (i) How is the performance of optimizing a min-max loss compared with that of optimizing the conventional pairwise surrogate loss based on mini-batch data?  It remains unclear which approach is faster and more robust for deep learning. (ii) How to optimize partial AUC for deep learning? Partial AUC  maximization is much more challenging than standard AUC maximization since the former involves the ordering of prediction scores among a large number of examples. (iii) How to optimize area under precision-recall curve (AUPRC)?  AUPRC is shown to be more appropriate for assessing the performance of a classifier on highly imbalanced data~\cite{davis06,10.1371/journal.pone.0118432}. Its close sibling performance metric named averaged precision (AP) and its optimization has attracted much attention in information retrieval and computer vision~\cite{brown2020smooth,NIPS2006_af44c4c5,Cakir_2019_CVPR, Chen_2019_CVPR,chen2020ap,10.5555/2984093.2984129,aistats17,Henderson_2017,Mohapatra2018EfficientOF,NEURIPS2020_b2eeb736,qin2008a,Rolinek_2020_CVPR}. Recently, there is a breakthrough on large-scale AP optimization with provable convergence guarantee by Qi et al.~\cite{DBLP:journals/corr/abs-2104-08736}. Nevertheless, it is still an open area for developing faster and robust methods for deep AUPRC maximization.

\textbf{Learning fair and interpretable AI models.} 
Building trustworthy AI is important for healthcare domains, in particular medical image classification. Two issues are of foremost importance, namely fairness~\cite{barocas-hardt-narayanan,DBLP:journals/corr/abs-1810-01943,DBLP:journals/corr/abs-2010-04053,DBLP:journals/corr/abs-1908-09635} and interpretability~\cite{chen2018looks,hase2019interpretable,arik2020protoattend}.  Although these issues have received tremendous attention in the literature for medical image classification~\cite{DBLP:journals/corr/abs-2102-06764,DBLP:journals/corr/abs-2003-00827,schutte2021using}, developing fair and interpretable DAM methods remains to be explored. 
%As AI is increasingly used in healthcare for decision making, it becomes more critical  to design AI models ensuring fairness for under-represented groups~\cite{barocas-hardt-narayanan,DBLP:journals/corr/abs-1810-01943,DBLP:journals/corr/abs-2010-04053,DBLP:journals/corr/abs-1908-09635} and to develop interpreatable AI models~\cite{}.  
%A number of fairness measures  and fairness-enhancing learning methods have been proposed in the literature, including multiple AUC-based fairness measures~\cite{borkan2019nuanced,kallus2019fairness}. %Although fairness issues have been considered in several recent works on medical image data~\cite{DBLP:journals/corr/abs-2102-06764,DBLP:journals/corr/abs-2003-00827}, it remain an undeveloped area for learning fair AI models under AUC-based fairness. 
Some outstanding questions and  work include %(i) which fairness is more appropriate than others in different scenarios; 
(i) how to develop scalable in-processing algorithms for optimizing AUC under AUC-based fairness constraints~\cite{borkan2019nuanced,kallus2019fairness}; (ii) how to develop scalable and interpretable DAM methods; (iii) evaluating these fairness-aware and interpretable AUC optimizaiton methods on large-scale medical image datasets.

\section{Conclusions}
In this extended abstract, we have discussed the history of AUC maximization in the last two decades. We then presented the recent studies on non-convex deep AUC maximization and its applications on various medical image classification problems. Finally, we discussed some outstanding challenges and new opportunities of deep AUC maximization, which serve as good research topics in the next few years. 

\bibliographystyle{plain}

\bibliography{AUC}

\end{document}